\documentclass[10pt, a4paper, conference, compsocconf]{IEEEtran}
\ifCLASSINFOpdf
\else
\fi

\usepackage{subcaption}
\usepackage{graphicx}
\usepackage{amsmath}
\usepackage{amsfonts}
\usepackage{stmaryrd}
\usepackage{tikz}
\usetikzlibrary[patterns,arrows,shapes]
\definecolor{lightgrey}{rgb}{0.73,0.73,0.73}
\definecolor{grey}{rgb}{0.38,0.38,0.38}

\begin{document}
%
\title{CNN training with graph-based sample preselection: application to handwritten character recognition}
%
\author{\IEEEauthorblockN{Fr\'{e}d\'{e}ric Rayar and Seiichi Uchida}
\IEEEauthorblockA{Department of Advanced Information Technology\\
Kyushu University, Fukuoka, Japan\\
\{rayar, uchida\}@human.ait.kyushu-u.ac.jp}
\and
\IEEEauthorblockN{Masanori Goto}
\IEEEauthorblockA{Research \& Development Center\\
 GLORY LTD., Hyogo, Japan\\
gotou.masanori@mail.glory.co.jp}}
\maketitle
%
\begin{abstract}
In this paper, we present a study on sample preselection in large training data set for CNN-based classification. To do so, we structure the input data set in a network representation, namely the Relative Neighbourhood Graph, and then extract some vectors of interest. The proposed preselection method is evaluated in the context of handwritten character recognition, by using two data sets, up to several hundred thousands of images. It is shown that the graph-based preselection can reduce the training data set without degrading the recognition accuracy of a non pre-trained CNN shallow model.
\end{abstract}

%

%
\IEEEpeerreviewmaketitle

\section{Introduction}
The advent of the so-called \textit{Deep Learning} in the last decade has led to major advances in several research fields such as artificial intelligence, pattern recognition, computer vision, and natural language processing. The document image analysis and recognition community has also embraced these neural network approaches for various tasks such as binarisation, layout analysis, character recognition, script identification, and word spotting. The number of paper that use such approaches is increasing rapidly, as illustrated by the recent ICDAR 2017 event\footnote{http://u-pat.org/ICDAR2017/program\_main.php}: 40 out of 52 oral papers in the main conference used neural approaches. Indeed, results obtained using these approaches are often impressive and almost always outperform state-of-the-art methods.

Among the existing architectures, Convolutional Neural Networks (CNN) have become a subject undergoing intense study in the past few years.
The usage of CNN falls into supervised machine learning, \textit{i.e.} one needs to gather a training data set and use this set to train a model, model that will then be used to predict various outputs. One specificity of using a neural approach is that one needs to have a huge amount of training data to perform well. Recently, the authors of~\cite{Uchida:2016} showed that increasing the size of the training data set allows to achieve a near-perfect recognition performance on handwritten digits. They have achieved up to 99.99\% of correct recognition using several hundred thousand training samples.

In this paper, we aim at studying the influence of a preselection step on the training data set, with regards to the recognition accuracy of a CNN-based classification. More precisely, we wonder if all the images of a large training data set are relevant in the training process. In what extent the underlying redundancy of such large training data set helps (or does not) during the training of a CNN model? Hence, we propose a graph-based preselection that removes samples lying around each class \textit{``center''}, without any severe degradation of the recognition performance.

The contributions of this paper are as follows:
\begin{enumerate}
	\item We propose a method for preselecting training images by analysing the data distribution. To do so, we structure the data in a network representation, namely the Relative Neighbourhood Graph (RNG)~\cite{Toussaint:1980}. Candidates are then extracted from this graph in order to be used in the CNN training process. 
	\item  We show through experimentations on two data sets, MNIST and HW\_O-RID (respectively 60,000 and 740,438 training images), that the proposed preselection strategy reduce the training data set up to 76\% without degrading the recognition accuracy.
\end{enumerate}

The rest of the paper is organised as follows: Section~\ref{sec:related} briefly presents the related works on data preselection in the literature. Section~\ref{sec:method} details the proposed preselection strategy using the RNG. In Section~\ref{sec:experiment}, we present the performed experiments to evaluate the relevance of our work and discuss the results in Section~\ref{sec:results}. Finally, we conclude our study in Section~\ref{sec:ccl} and outline directions for future research.\\
	
\section{Related Work}
\label{sec:related}

%

Training a classification model is often performed on large training data sets, to avoid overfitting and enhance the generalisation performance of the model. However, as mentioned in~\cite{Jankowski:2004}, several reasons can support the need of reducing the training set: (i) reducing the noise, (ii) reducing storage and memory requirement and (iii) reducing the computation time of the training or the prediction phase.

Many sample (or prototype) selection solutions have been proposed in the past to address training set reduction. We can categorise these solutions in three families:
\begin{enumerate}
	\item The \textit{``editing''} paradigm, that aims at eliminating erroneous instances and remove possible overlapping between classes. Hence, such algorithms behave as noise filters. For instance, the edited nearest neighbour~\cite{Wilson:1972} algorithm removes an instance if its class is inconsistent with its neighbours' majority class.
	\item The \textit{``condensing''} paradigm, that aims at finding instances that will allow to perform as well as a Nearest Neighbour (NN) classifier that uses the whole training set. For instance, the condensed nearest neighbour rule~\cite{Hart:1968}, removes instances from the training set one by one if their absence do not degrade the classification accuracy. However, as mentioned in~\cite{Jankowski:2004}, such techniques are \textit{``very fragile in respect to noise and the order of presentation''}.
	\item The \textit{``hybrid''} (editing-condensing) paradigm, that aims at removing noise and redundant instances at the same time.
\end{enumerate}

Among the proposed techniques, that falls into the aforementioned categories, it is worth mentioning the usage of: (i) random selection techniques~\cite{Lee:2007}, (ii) clustering techniques~\cite{Tran:2003} or graph-based techniques~\cite{Toussaint:1985}. One can refer to thorough surveys that have been done recently by Garcia et al.~\cite{Garcia:2012} in 2012 (for NN based classification), and by Jung et al.~\cite{Jung:2014} in 2014 for (Support Vector Machine (SVM)~\cite{Corte:1995} based classification). 

As one can deduce by the existence of the aforementioned surveys, prototype selection has been widely studied for the NN-based classification and SVMs, but to the best of our knowledge, no similar studies, has not been performed for CNN (or more generally neural networks). Indeed, most of the studies that use CNN are usually focused on the acquirement of large training data sets, using crowdsourcing (e.g. ImageNet~\cite{ImageNet:2009}), synthetic data generation or data augmentation~\cite{Wong:2016} techniques.

In this study, we have chosen a graph-based condensing sample preselection strategy. Indeed, an extensive series of work have been performed on the relevance of proximity graphs~\cite{Toussaint:1991} to preselect samples in classification training set. 
More specifically, we have selected the RNG that has been recently proven a good fit to preselect high-dimensional samples~\cite{Toussaint:2012} in large training data sets~\cite{Goto:2015}.\\
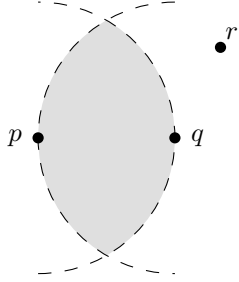
\begin{figure}[!t]
\centering
\begin{tikzpicture}[x=1.0cm,y=1.0cm,scale=0.6]
	\begin{scope}[draw=white]
	\draw [clip] (10.0,-3.0) arc (-90:90:3cm);
	\fill [color=lightgray, fill opacity=0.5] (13.0,3.0) arc (90:270:3cm);
	\end{scope}
	
	\draw [dash pattern=on 5pt off 5pt, pattern color=black] (10.0,-3.0) arc (-90:90:3cm);
	\draw [dash pattern=on 5pt off 5pt, pattern color=black] (13.0,3.0) arc (90:270:3cm);
	\fill [color=black] (10.0,0.0) circle (3.5pt);	
	\draw[color=black] (9.50,0.0) node {$p$};		
	\fill [color=black] (13.0,0.0) circle (3.5pt);	
	\draw[color=black] (13.50,0.0) node {$q$};		
	\fill [color=black] (14.0,2.0) circle (3.5pt);	
	\draw[color=black] (14.25,2.3) node {$r$};		
\end{tikzpicture}
\caption{Relative neighbourhood (grey area) of two points $p, q \in \mathbb{R}^2$. If no other point of $D$ lays in this neighbourhood, then $p$ and $q$ are relative neighbours.}
\label{fig:rng}
\end{figure}

\section{Preselection using the Relative Neighbourhood Graph}
\label{sec:method}

\subsection{Relative Neighbourhood Graph}
\label{subsec:rng}

The relative neighbourhood graph has been introduced by G. Toussaint in the early 1980s~\cite{Toussaint:1980}. The construction of this graph is based on the notion of \textit{``relatively closeness''}, that defines two points as relative neighbours if ``\textit{they are at least as close to each other as they are to any other points}''. From this definition, one can define ${\rm RNG}=(V,E)$ as the graph built from a given data set$D$, where distinct points $p$ and $q$ of $D$ are connected by an edge $\overline{pq}$ if and only if they are relative neighbours. Thus,
\begin{gather*}
E({\rm RNG}) = \{ \overline{pq}\text{ } |\text{ }  p,q \in D, p \neq q, \\ 
\delta(p,q) \leq \max(\delta(p,r),\delta(q,r)), \forall r \in D\backslash\{p,q\} \}.
\end{gather*}
where $\delta: D \times D \rightarrow \mathbb{R}$ is a distance function.
An illustration of the relative neighbourhood of two points $p, q \in \mathbb{R}^2$ is given in Figure \ref{fig:rng}.

\begin{figure}[!t]
	\centering
	\includegraphics[width=0.6\linewidth]{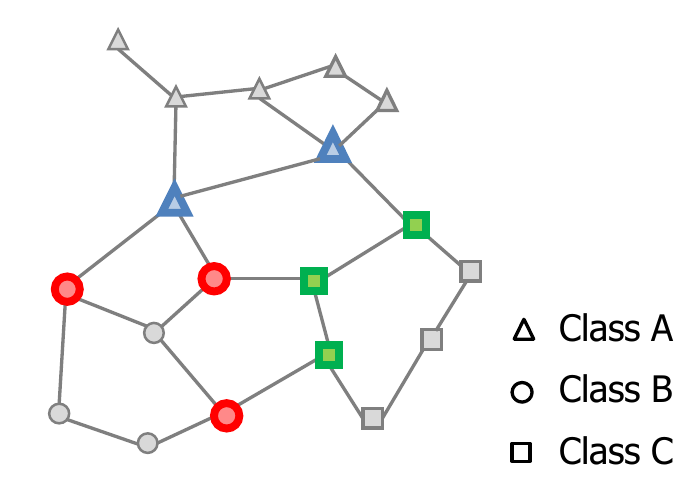}%
	\caption{Illustration of bridge vectors on a toy data set. The bridge vectors are highlighted with colours and thicker borders. }
	\label{fig:bv}
\end{figure}

The RNG has been used in several works in various fields such as computer vision, geographic analysis or pattern classification~\cite{Toussaint:2014}. Its main benefit is that it is a connected graph, that highlights the topology of the data and embeds local information about vertex neighbourhood.
The main drawback of the RNG is its construction complexity in $O(|D|^3)$. However, recent works such as~\cite{Rayar:2015b} and~\cite{Goto:2015} have proposed solutions to build RNG of large data sets, up to millions data points.

\subsection{Bridge Vectors}
\label{subsec:bv}

Bridge vectors, as defined in~\cite{Goto:2015}, are points that lay in the outer frontiers of classes. Using the RNG representation of the data set, they correspond to points that have at least one relative neighbour from a different class. Figure~\ref{fig:bv} illustrates the notion of bridge vectors on a toy data set.

In our experiments, to get the bridge vectors of a given training data set, the following straightforward steps are performed: (i) build the RNG of the training data set using the algorithm proposed in~\cite{Goto:2015} and (ii) if two points of different classes are connected by an edge in the computed graph, they are added in the bridge vectors' list.

\begin{figure*}[!t]
	\centering
	\includegraphics[width=0.7\linewidth]{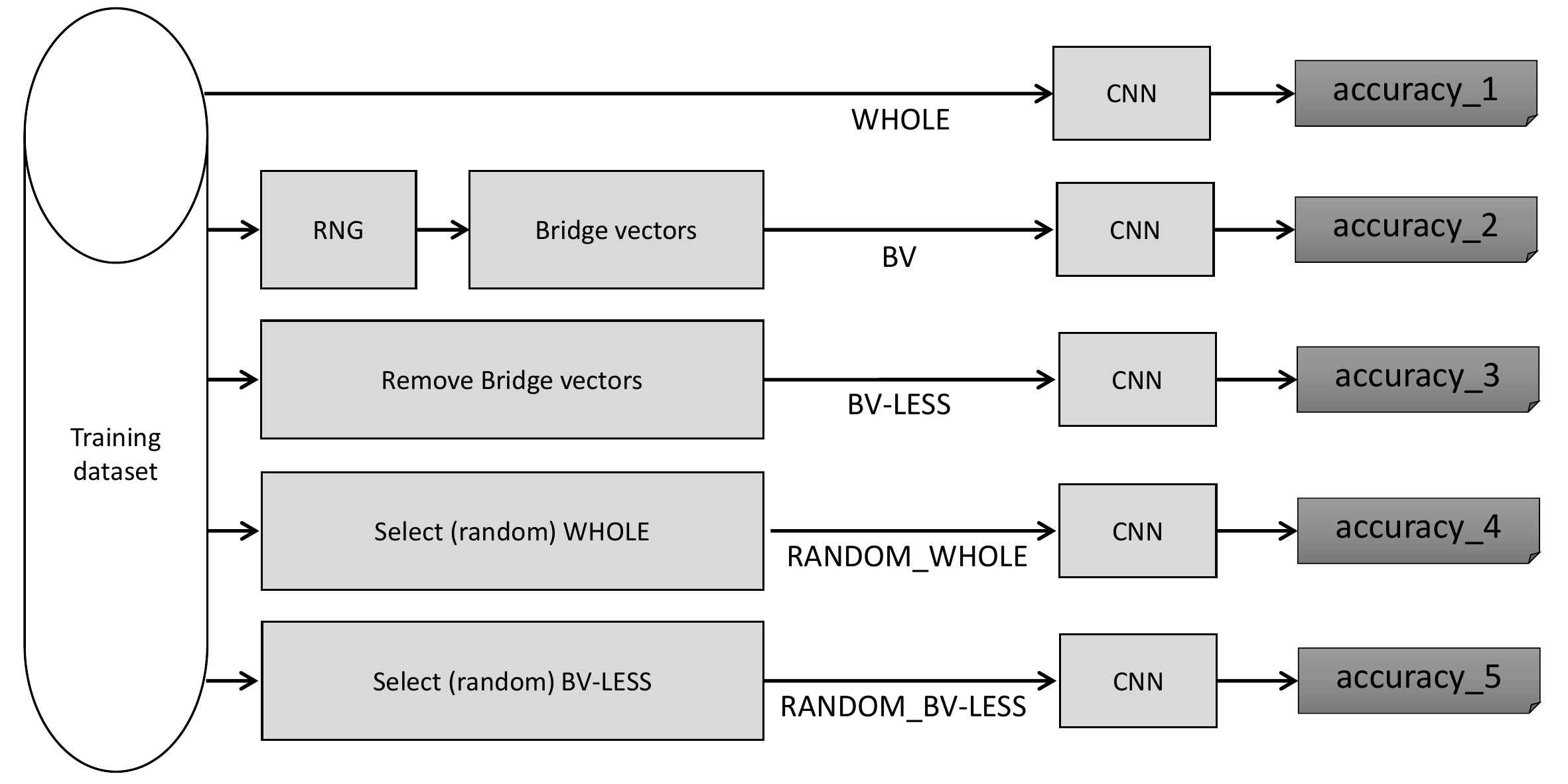}%
	\caption{General workflow of the study. The CNN model is trained using either: (1) $\rm{WHOLE}$, (2) $\rm{BV}$, (3) $\rm{BV{\text -}LESS}$, (4) $\rm{RANDOM\_WHOLE}$ or (5)  $\rm{RANDOM\_BV{\text -}LESS}$, as defined in Section~\ref{subsec:workflow}.}
	\label{fig:workflow}
\end{figure*}

\subsection{Study workflow}
\label{subsec:workflow}

Figure~\ref{fig:workflow} illustrates the workflow of the proposed study. The main goal is to evaluate the relevance of the proposed preselection technique, and highlight the importance of bridge vectors in the training of a CNN classifier. 

To do so, five different training subsets have been used for a given data set:
\begin{itemize}
	\item $\rm{WHOLE}$: the whole training data set,
	\item $\rm{BV}$: only the extracted bridge vectors of the RNG build from $\rm{WHOLE}$,
	\item $\rm{BV{\text -}LESS}$: the data set $\rm{WHOLE}$, but without bridge vectors $\rm{BV}$,
	\item $\rm{RANDOM\_WHOLE}$: a random subset of $\rm{WHOLE}$, with approximatively the same size as $\rm{BV}$,
	\item $\rm{RANDOM\_BV{\text -}LESS}$: a random subset of $\rm{BV{\text -}LESS}$, with approximatively the same size as $\rm{BV}$.
\end{itemize}

\section{Experimental setup}
\label{sec:experiment}

\subsection{Data sets}
\label{subsec:dataset}

We applied the proposed training data set preselection to two isolated handwritten digit data sets, namely MNIST and HW\_R-OID. Table~\ref{tab:data set} shows the size of each subset, for the two data sets that have been used in this work.

First, the MNIST~\cite{Lecun:1998} data set, that corresponds to binary 28~$\times$~28 images of centered handwritten digits. Ground truth (\textit{i.e.} correct class label (\textit{``0''}, $\dots$, \textit{``9''}), is provided for each image. In our experiments, we have used 60,000 images in the training data set and 10,000 for testing purpose.

Second, the HW\_R-OID data set is an original data set from~\cite{Uchida:2016}. It contains 822,714 images collected from forms written by multiple people. The images are 32~$\times~32$ binary images of isolated digits and ground-truth is also available. In this data set, the number of the samples of each class is different but almost the same (between 65,000 and 85,000 samples per class, except the class \textit{``0''} that has slightly more than 187,000 samples). In our experiments, we have split the data set in train/test subsets with a 90/10 ratio (740,438 training + 82,276 test images). To do so, 90\% of each class samples have been gathered to build the training subset.

For the these two data sets, the intensities of the raw pixels have been used to described the images, and the Euclidean distance has been used to compute the similarity between pairs of images.

\subsection{CNN classification}
\label{subsec:cnn}

Experiments were done on a computer with a i7-6850K CPU @3.60GHz, with 64.0GB of RAM (not all of it was used during runtime), and a NVIDIA GeForce GTX 1080 GPU. Our CNN classification implementation relies on the usage of Python (3.6.2) along with the Keras library (2.0.6) and a TensorFlow (1.3.0) backend.
	
\begin{figure}[!b]
	\centering
	\includegraphics[width=0.9\linewidth]{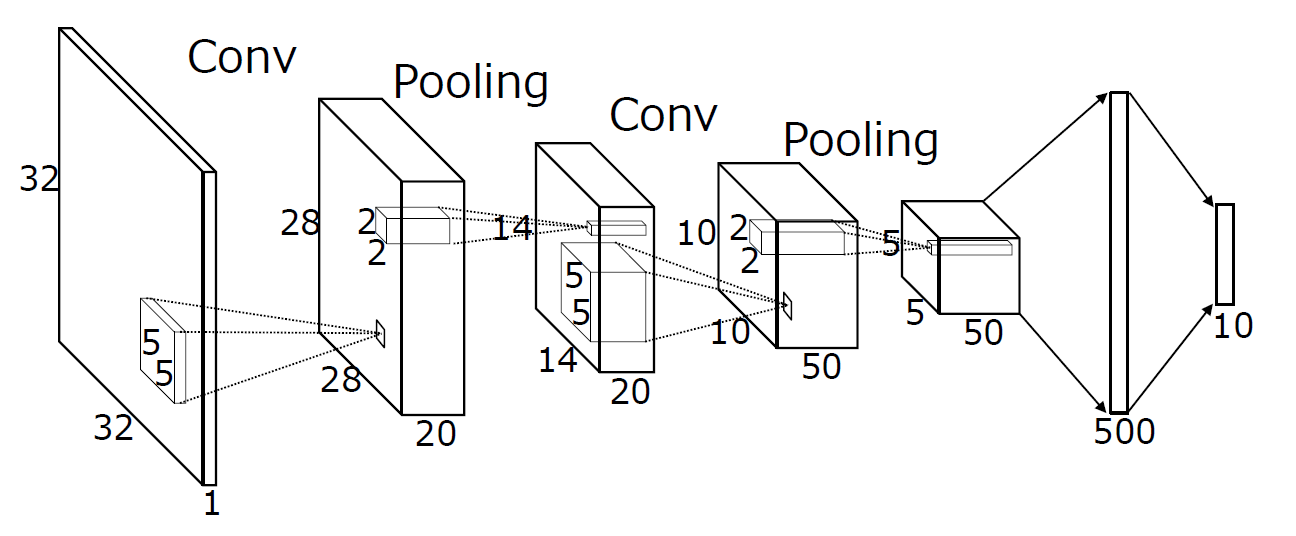}%
	\caption{Modified LeNet-5 CNN architecture used in our experiments. No pretraining has been done. Inputs images are $28 \times 28$ and $32 \times 32$ for the MNIST and HW\_R-OID data sets respectively.}
	\label{fig:cnn}
\end{figure}

The same CNN structure and parameters as~\cite{Uchida:2016} have been used. A simple CNN architecture, namely modified LeNet-5. The main difference with the original LeNet-5~\cite{Lecun:1998} is the usage of ReLU and max-pooling functions for the CONV layers. Figure~\ref{fig:cnn} illustrates the architecture that has been used. As mentioned in~\cite{Uchida:2016}, it is \textit{``a rather shallow CNN compared to the recent CNNs. However, it still performed with an almost perfect recognition accuracy''} (when trained with a large data set).
No pre-initialisation of the weights is done, and the CNN is trained with classical back-propagation for 10 epochs.

\begin{table*}[!t]
\renewcommand{\arraystretch}{1.3}
\caption{CNN-based classification results: (i) size of the training data set, (ii) average recognition accuracy and (iii) average training time (in seconds) are presented.}
\label{tab:data set}
\centering
	\begin{tabular}{|c|c|c|c|c|c|c|}
		\hline
		      & Training data set& $\rm{WHOLE}$ & $\rm{BV}$ & $\rm{BV{\text -}LESS}$ & $\rm{RANDOM}$ & $\rm{RANDOM}$ \\
		      & &         &      &                   & $\rm{WHOLE}$  & $\rm{BV{\text -}LESS}$ \\
		\hline
		\hline
		      & \# training data  & 60,000  & 22,257  & 37,743  & 22,260  & 22,258  \\
		      \cline{2-7}
		MNIST & accuracy (\%)     & 98.79 & 98.78 & 97.05 & 98.22 & 96.47 \\
			\cline{2-7}
		      & training time (s) & 252     & 104     & 164     & 104     & 103     \\
		\hline
		\hline
		           & \# training data   & 740,438   & 173,808   & 566,630   & 174,002   & 174,012   \\
		 \cline{2-7}
		 HW\_R-OID & accuracy (\%)      & 99.9343 & 99.9314 & 99.7372 & 99.8586 & 99.5631 \\
		 \cline{2-7}
		           & training time (s)  & 4171      & 1086      & 3252      & 1086      & 1089       \\
		\hline
	\end{tabular}
\end{table*}

\vspace{3mm}

\begin{figure*}[!th]
	\centering
	\includegraphics[width=0.48\linewidth]{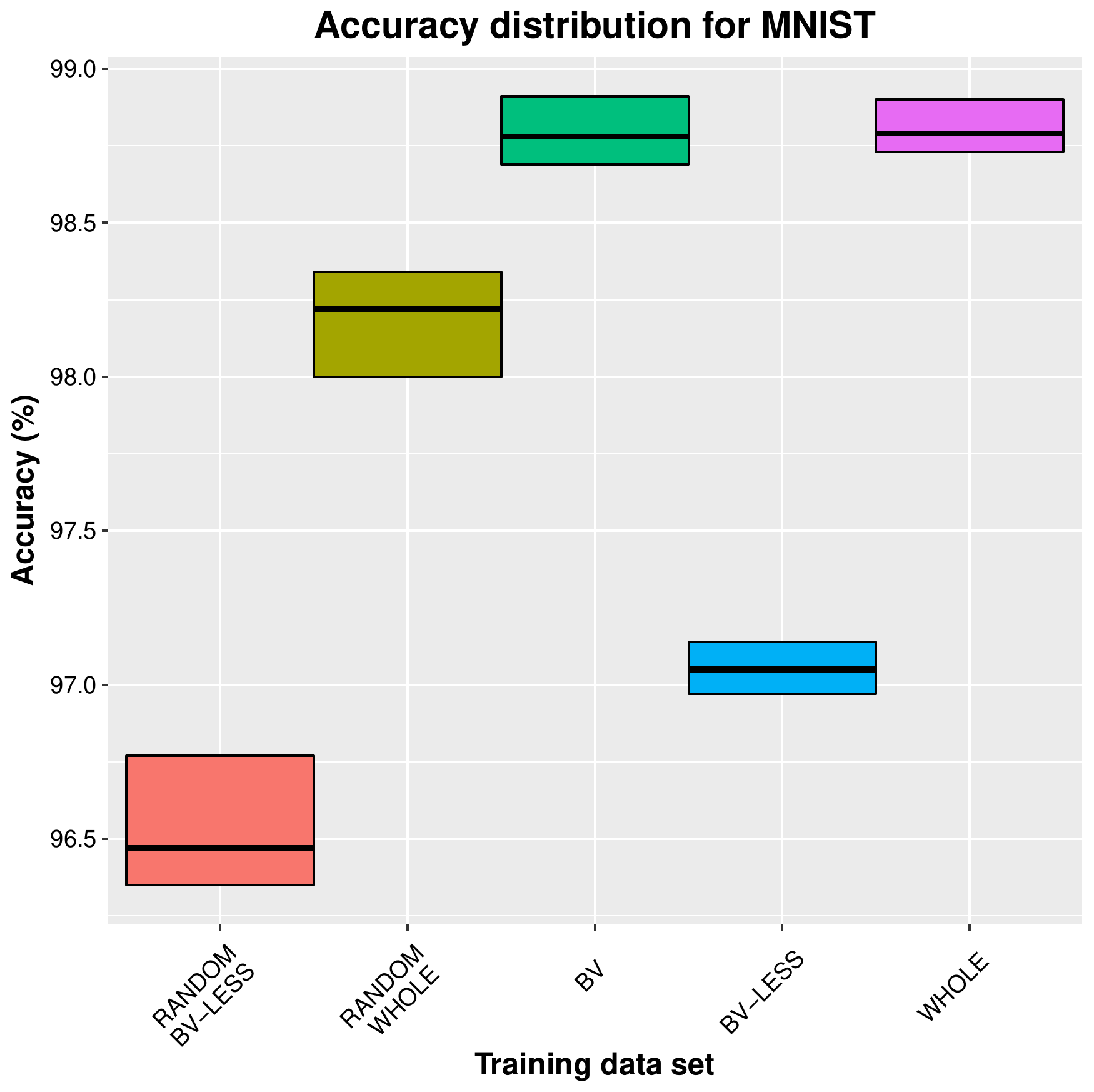}%
	\hspace{5mm}
	\includegraphics[width=0.48\linewidth]{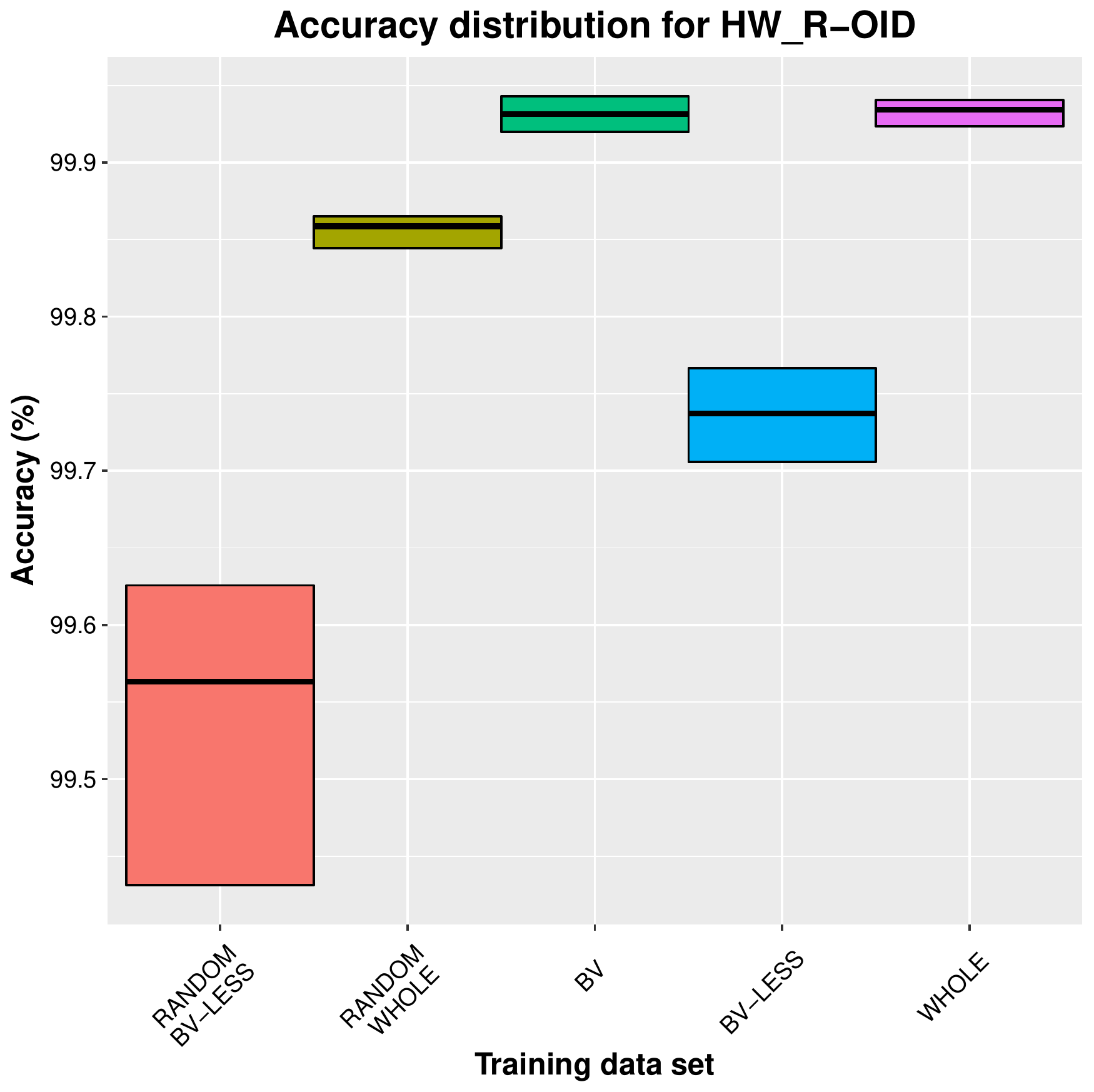}%
	\caption{Accuracy distribution on MNIST (left) and HW\_R-OID (right) data sets over the different training sets (presented in ascending cardinality).}
	\label{fig:exp_acc}
\end{figure*}

During our experimentations, both computation times and recognition accuracies have been measured for further analysis. For each training data sets, experiments were run 5 times to compute an average value of the aforementioned metrics.\\

\section{Results}
\label{sec:results}

\subsection{Analysis of the preselection efficiency}
\label{subsec:exp_acc}

Table~\ref{tab:data set} presents the average accuracies obtained for all the training data sets introduced in Section~\ref{subsec:workflow} for both MNIST and HW\_R-OID. The major observation is that the average accuracy obtained by training the CNN with only the bridge vectors ($\rm{BV}$) is almost the same as the one obtained by training the CNN with the whole available training data set ($\rm{WHOLE}$). This supports the efficiency of the proposed preselection strategy.

By removing the bridge vectors from the training data set, the CNN recognition performance is lower, while the cardinality of the $\rm{BV{\text -}LESS}$ training data set remains greater than the bridge vector subset ($\rm{BV}$). Hence, the size of the training data set is not the only criterion that should be considered while training a CNN.

To emphasize on the relevance of the proposed preselected candidates, we also have randomly selected samples from $\rm{WHOLE}$ and $\rm{BV{\text -}LESS}$ data sets (to produce $\rm{RANDOM\_WHOLE}$ and $\rm{RANDOM\_BV{\text -}LESS}$ respectively). The number of samples has been selected close to the the cardinality of the number of bridge vectors.  The $\rm{RANDOM\_WHOLE}$ training data set allows the CNN to perform better than both the $\rm{BV{\text -}LESS}$ and the $\rm{RANDOM\_BV{\text -}LESS}$. Hence, the sole presence of bridge vectors in the training data set seems to allow the CNN to achieves better recognition accuracy.

To avoid basing our conclusions only on average accuracies, we also present the accuracy distribution obtained for each training data sets (in ascending cardinality) over the 5 runs in Figure~\ref{fig:exp_acc}. It clearly helps us to assert the following statement: in terms of recognition accuracy, we have: 
\begin{gather*}
	  \rm{WHOLE} \approx \rm{BV} > \rm{RANDOM\_WHOLE} > \dots \\
	 \dots \text{ } \rm{BV{\text -}LESS} > \rm{RANDOM\_BV{\text -}LESS}
\end{gather*}

Hence, we validate our hypothesis that not all the training images are mandatory to achieve high recognition accuracy. Some redundancy in the training data set can be pruned using the proposed preselection strategy, while allowing the CNN to achieve the same near-perfect recognition accuracies.

\subsection{Analysis of the method overhead}
\label{subsec:exp_time}

Table~\ref{tab:time} presents the computation times for the network representation of the training data set, using the RNG construction algorithm of~\cite{Goto:2015}. Since the algorithm takes the images as input, we also provide the data loading time.
For the smallest data set, only $5$ minutes are needed to generate the RNG. For the larger data set, this computation time increases: about $17$ hours. Hence, the computation of the RNG may be a limitation, even if this could be addressed by different solutions (\textit{e.g.} the optimisation of the code or the usage of a GPU).

Nonetheless, since the computation of the RNG, for a given training data set, is performed only once, the ratio of this overhead can be minimised in a general CNN training framework, where one needs to make several number of trial-and-error iterations regarding the choice of the neural network structure and parameters.


Table~\ref{tab:data set} presents the training subsets' size and the average training times of our experiments. Regarding the CNN training, thanks to the proposed preselection technique, we can decrease the training data set by $76\%$ and $63\%$, while having a training speed-up ratio of  $3.8$ and $2.4$ for the MNIST and HW\_R-OID data sets respectively. 

\begin{table}[!t]
\renewcommand{\arraystretch}{1.3}
\caption{Image loading time and RNG computation time (in seconds).}
\label{tab:time}
\centering
	\begin{tabular}{|c|c|c|c|}
		\hline
		      			& \# training & image load 	& RNG computation \\
		      			& images      & loading (s) & computation (s)\\
		\hline
		\hline
		      MNIST		& 60,000	& 133  	      	& 304 \\
		\hline
		      HW\_R-OID	& 740,438	& 1397			& 61,270 \\
		\hline		      
	\end{tabular}
\end{table}

\subsection{Analysis of bridge vectors}
\label{subsec:exp_bv}

\begin{table*}[!th]
\renewcommand{\arraystretch}{1.3}
\caption{For a given (while) training data set: (i) the first two rows correspond to the number of data per digit and the ratio over the training set (in ); (ii) the next rows correspond to the number of bridge vectors per digit, the ratio over the class elements and the ratio over the whole bridge vector set.  Percentage values are rounded off to two decimal places.}
\label{tab:analysis}
\centering
	\begin{tabular}{|c|c|c|c|c|c|c|c|c|c|c|c|}
		\hline
		      	& class & $0$ & $1$ & $2$ & $3$ & $4$ & $5$ & $6$ & $7$ & $8$ & $9$\\
		\hline
		\hline
		       & \# data &  5923 & 6742 & 5958 & 6131 & 5842 & 5421 & 5918 & 6265 & 5851 & 5949\\
			    & class () & 9.87 & 11.24 & 9.93 & 10.22 & 9.74 & 9.03 & 9.87 & 10.44 & 9.75 & 9.91 \\

		\cline{2-12}
		       MNIST & \# BV& 1518 & 828 & 2113 & 2917 & 2498 & 2794 & 1375 & 2021 & 2864 & 3329\\
			   & class (\%) & 25.6 & 12.3 & 35.46 & 47.58 & 42.76 & 51.54 & 23.23 & 32.26 & 48.95 & 55.96 \\	       
			   & BV (\%) & 6.82 & 3.72 & 9.49 & 13.10 & 11.22 & 12.55 & 6.18 & 9.08 & 12.87 & 14.95 \\
		\hline       
		\hline
		      	& \# data &  168,521 & 58,202 & 76,987 & 65,848 & 60,030 & 60,075 & 58,172 & 56,723 & 73,176 & 62,704\\
			 & class (\%) & 22.76 & 7.86 & 10.39 & 8.89 & 8.10 & 8.11 & 7.85 & 7.66 & 9.88 & 8.47 \\

		\cline{2-12}
		      HW\_R-OID & \# BV& 13,728 & 9,204 & 10,581 & 19,682 & 16,054 & 17,233 & 12,636 & 19,983 & 28,241 & 26,466\\
			   & class (\%) & 8.15 & 15.81 & 13.74 & 29.89 & 26.74 & 28.68 & 21.72 & 35.23 & 38.59 & 42.20 \\	       
			   & BV (\%) & 7.90 & 5.29 &6.08 & 11.32 & 9.24 & 9.91 & 7.27 & 11.49 & 16.25 & 15.23 \\
		\hline
	\end{tabular}
\end{table*}

\begin{figure*}[!t]
	\centering
	\includegraphics[width=0.49\linewidth]{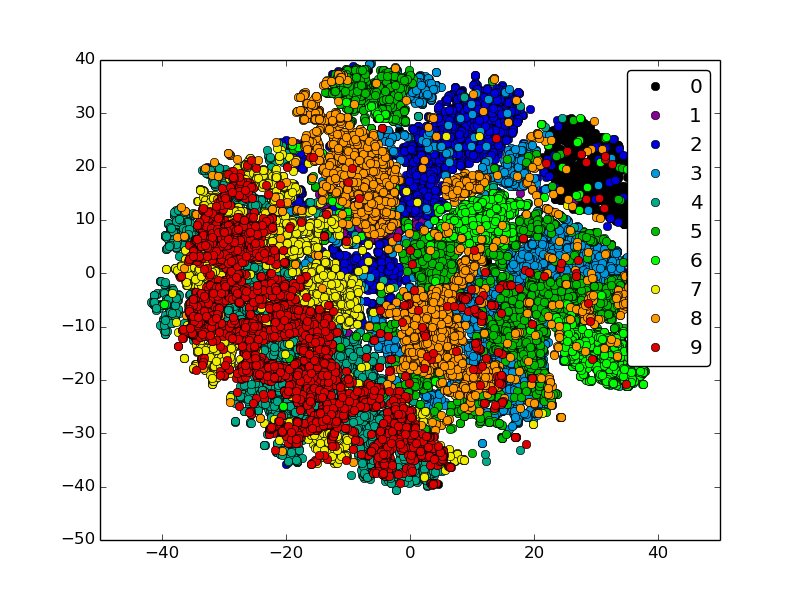}%
	\includegraphics[width=0.49\linewidth]{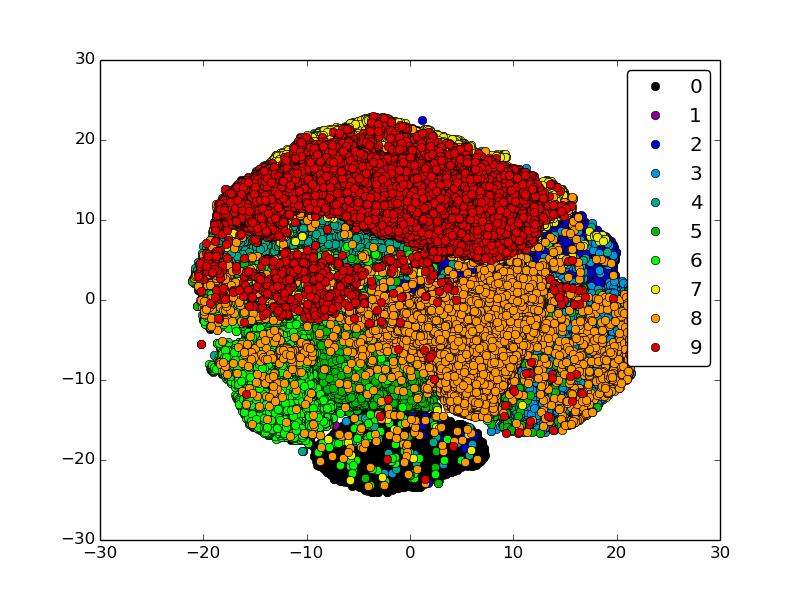}%
	\caption{t-SNE visualisation of the bridge vectors of MNIST (left) and HW\_R-OID (right) data sets (using $\rm{perplexity}=30.0$).}
	\label{fig:exp_tsne}
\end{figure*}

Figures~\ref{fig:exp_bvex_mnist} and~\ref{fig:exp_bvex_hw}, provided in Appendix~\ref{sec:appendix_bvex} illustrate some bridge vectors, extracted from the whole training data set of MNIST and HW\_R-OID respectively. One can see that while some ambiguous instances are present (last two columns, manually selected), the major part of the bridge vectors are \textit{``standard''} patterns, that allow the CNN to recognize such patterns well during the recognition step.

Table~\ref{tab:analysis} presents the quantitative results of the bridge vectors analysis for the two studied data sets. The figures provided in Appendix~\ref{sec:appendix_bv} present visual aids to allow one to have a better understanding of the data sets composition.

In the MNIST data set, the classes in the WHOLE training set are well balanced. One can note that class \textit{``1''} is the class that had the most number of images in the training  data set. However, it is the class that has the smallest number of bridge vectors (less than $4\%$). 

A similar observation can also be made for the HW\_R-OID data set: indeed, class \textit{``0''} has at least two times more samples than the other classes. However, only $8.15\%$ of class \textit{``0''} samples are bridge vectors, which represent only $7.90\%$ of all the bridge vectors. Hence, one can deduce that for a given class, if most of the samples are very similar, they do not make a huge contribution during the CNN model training.

A second observation that can be made is that for both data sets, certain classes have more bridge vectors (more than $10\%$ of the BV subset): \{\textit{``3''},\textit{``4''},\textit{``8''},\textit{``9''}\}  and \{\textit{``8''},\textit{``9''}\} for MNIST and HW\_R-OID respectively. This observation is consistent with the similarity of such handwritten digits observed in the data sets. 

To visualise the separability of the bridge vectors, the t-distributed stochastic neighbour embedding (t-SNE)~\cite{vanDerMaaten:2008} algorithm has been used. Several values of $\rm{perplexity}$ have been used, within the interval $ [5, 50]$ as suggested in~\cite{vanDerMaaten:2008}, and we observed the same type of results.

Figure~\ref{fig:exp_tsne} presents the t-SNE visualisations, using $perplexity=30.0$, of the bridge vectors for the two studied data sets.
For MNIST, classes \textit{``4''} (in turquoise) and \textit{``9''} (in red) are clearly not separated. For the second data set, one can note that classes \textit{``8''} (in orange) and \textit{``9''} (in red) are well-separated but overlap with several other classes.

\subsection{Residual error analysis}
\label{subsec:exp_error}

We have performed a residual error analysis following the CNN-based classification on both MNIST and HW\_R-ID data sets. Indeed, since the proposed preselection method do not degrade the recognition accuracies, the following question arises: what is the relation(s) between misclassified digit sets using either $\rm{WHOLE}$, $\rm{BV}$ or $\rm{BV{\text -}LESS}$ as training set. 

The figures~(\ref{fig:_exp_mnist_4}),~(\ref{fig:_exp_hw_4}),~(\ref{fig:_exp_mnist_5}) and~(\ref{fig:_exp_hw_5}), provided in Appendix~\ref{sec:appendix_cnn}, present visual aids to allow one to have a better understanding of misclassified set relations. Regarding the misclassified sets of $\rm{WHOLE}$ and $\rm{BV}$, it is interesting to note that the major part of misclassification obtained using only the bridge vectors are also misclassified using the whole training dataset (they correspond to the intersection between the two sets). 

Figure~\ref{fig:exp_mis} illustrates all the misclassified (sorted) digits, using either $\rm{WHOLE}$ or $\rm{BV}$ as training data set for HW\_R-OID. One can easily spot some rather difficult instances of handwritten digits, even for a human being. However, the CNN still fails to classify some instances that are ``easy'' for a human (\textit{e.g.} the last instances of \textit{``9''} using only the bridge vectors).

Regarding the misclassified sets of $\rm{BV}$ and $\rm{BV{\text -}LESS}$, one can see that  the presence of the bridge vectors in the training set allows the CNN to correctly classify more instances, reducing drastically the number of misclassification ($-60\%$ and $-71\%$ for MNIST and HW\_R-OID respectively).

\section{Conclusion}
\label{sec:ccl}

We have provided in this study some insights about the relevance of the training data that is fed to train a CNN model. Indeed, even if a huge quantity of training samples helps to achieve almost near-perfect recognition, its underlying redundancy does not have a huge impact on the model's training. The proposed graph-based preselection method allows to reduce the training data set considerably, and thus accelerates the training computation time without deteriorating the recognition accuracy.

Future works will be done towards the following directions: (i) address the limitation related to the RNG computation time, (ii) perform experimentations using data sets other than isolated handwritten characters and (iii) start a formal study on the existence of \textit{``support vectors''} for CNN.

\section*{Acknowledgement}
This research was partially supported by MEXT-Japan (Grant No. 17H06100).


\bibliographystyle{IEEEtran}
\bibliography{arxiv}

\clearpage
\newpage

\onecolumn

\appendices

\section{Samples of bridge vectors}
\label{sec:appendix_bvex}

\begin{figure}[!h]
	\centering
	\includegraphics[width=0.55\linewidth]{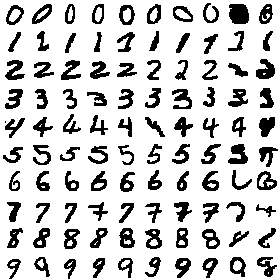}%
	\caption{MNIST bridge vector examples, extracted from the whole training data set. The two last columns show ambiguous instances, that have been manually selected.}
	\label{fig:exp_bvex_mnist}
\end{figure}

\vspace{5mm}

\begin{figure}[!h]
	\centering
	\includegraphics[width=0.55\linewidth]{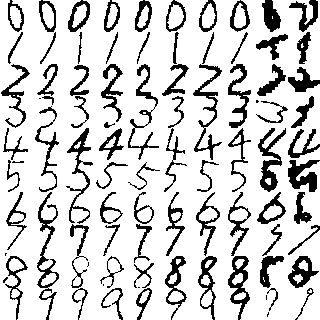}%
	\caption{HW\_R-OID bridge vector examples, extracted from the whole training data set. The two last columns show ambiguous instances, that have been manually selected.}
	\label{fig:exp_bvex_hw}
\end{figure}

\section{Analysis of bridge vectors - Visual aids}
\label{sec:appendix_bv}

\begin{figure*}[!h]
    \centering
    \begin{subfigure}[t]{0.49\textwidth}
    	\centering
		\includegraphics[width=1.0\linewidth,viewport=50bp 240bp 550bp 570bp,clip]{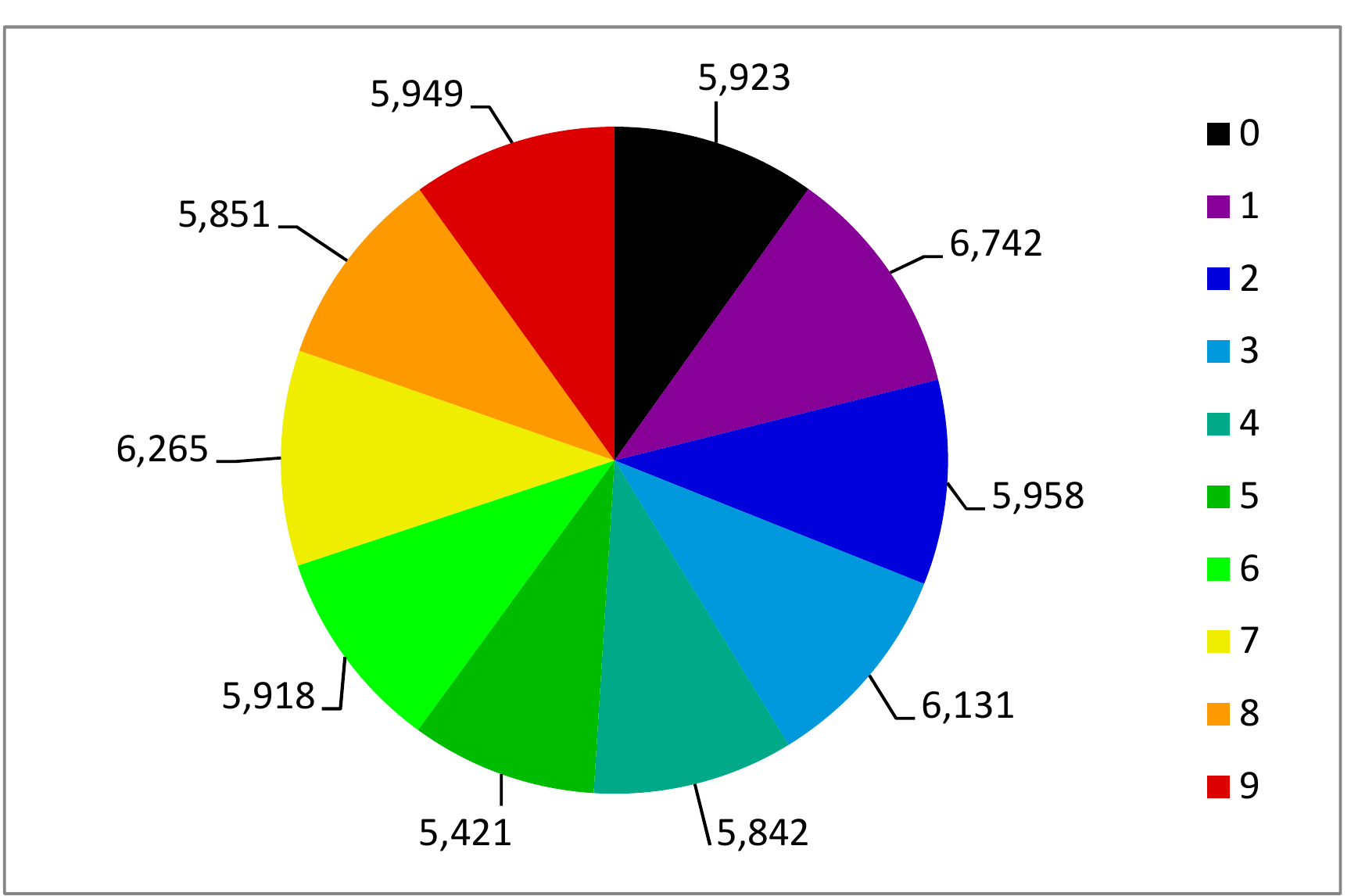}%
		\caption{MNIST training dataset: class distribution.}
		\label{fig:_exp_mnist_1}
    \end{subfigure}%
    ~ 
    \begin{subfigure}[t]{0.49\textwidth}
        \centering
		\includegraphics[width=1.0\linewidth,viewport=50bp 240bp 550bp 570bp,clip]{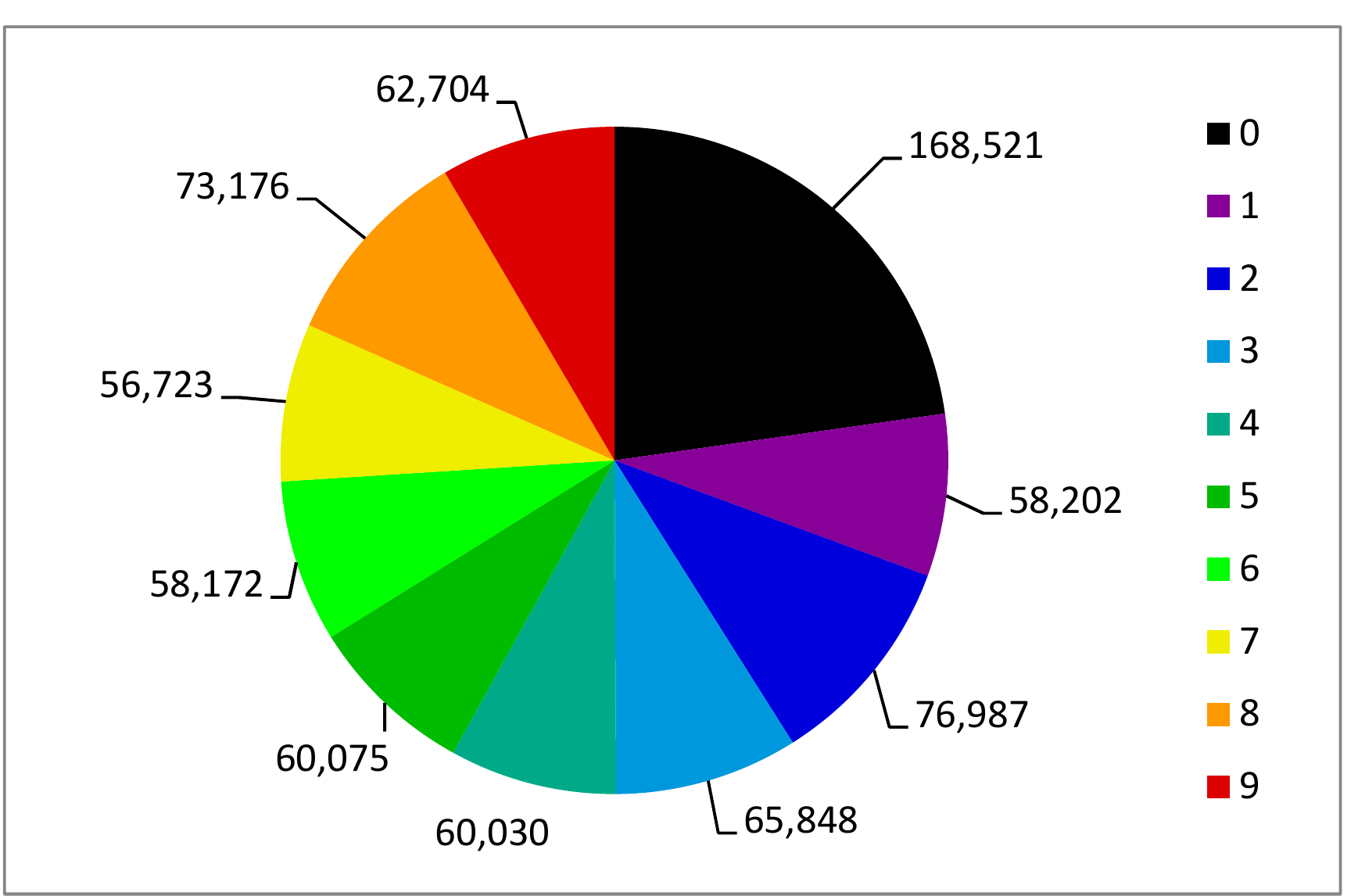}%
		\caption{HW\_R-OID training dataset: class distribution.}
		\label{fig:_exp_hw_1}
	\end{subfigure}%
 
	\vspace{10mm}
	
	\begin{subfigure}[t]{0.49\textwidth}
		\centering
		\includegraphics[width=1.0\linewidth,viewport=50bp 240bp 550bp 570bp,clip]{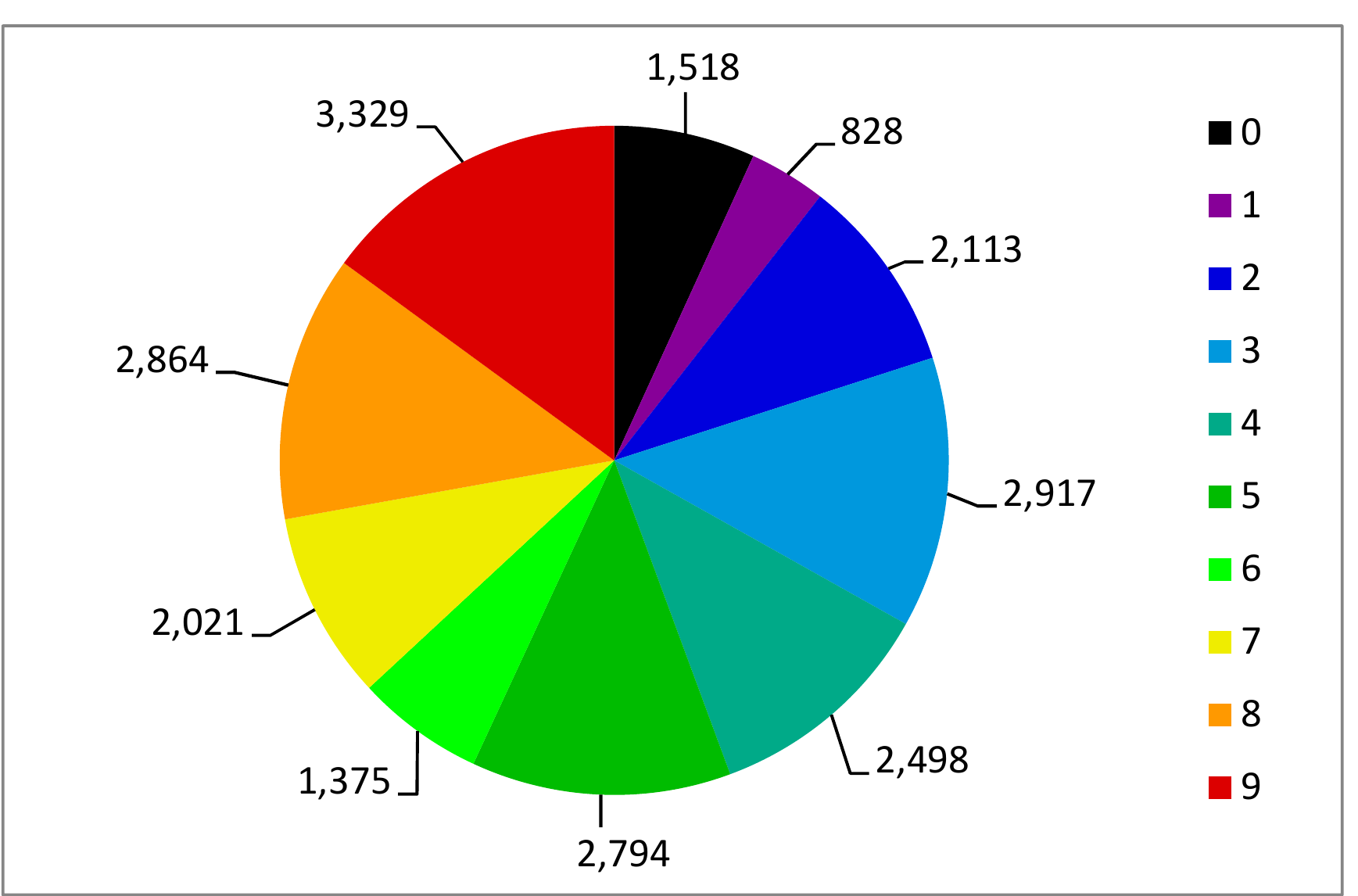}%
		\caption{MNIST training dataset: bridge vector distribution.}
		\label{fig:_exp_mnist_2}
	\end{subfigure}%
    ~ 
	\begin{subfigure}[t]{0.49\textwidth}
		\centering
		\includegraphics[width=1.0\linewidth,viewport=50bp 240bp 550bp 570bp,clip]{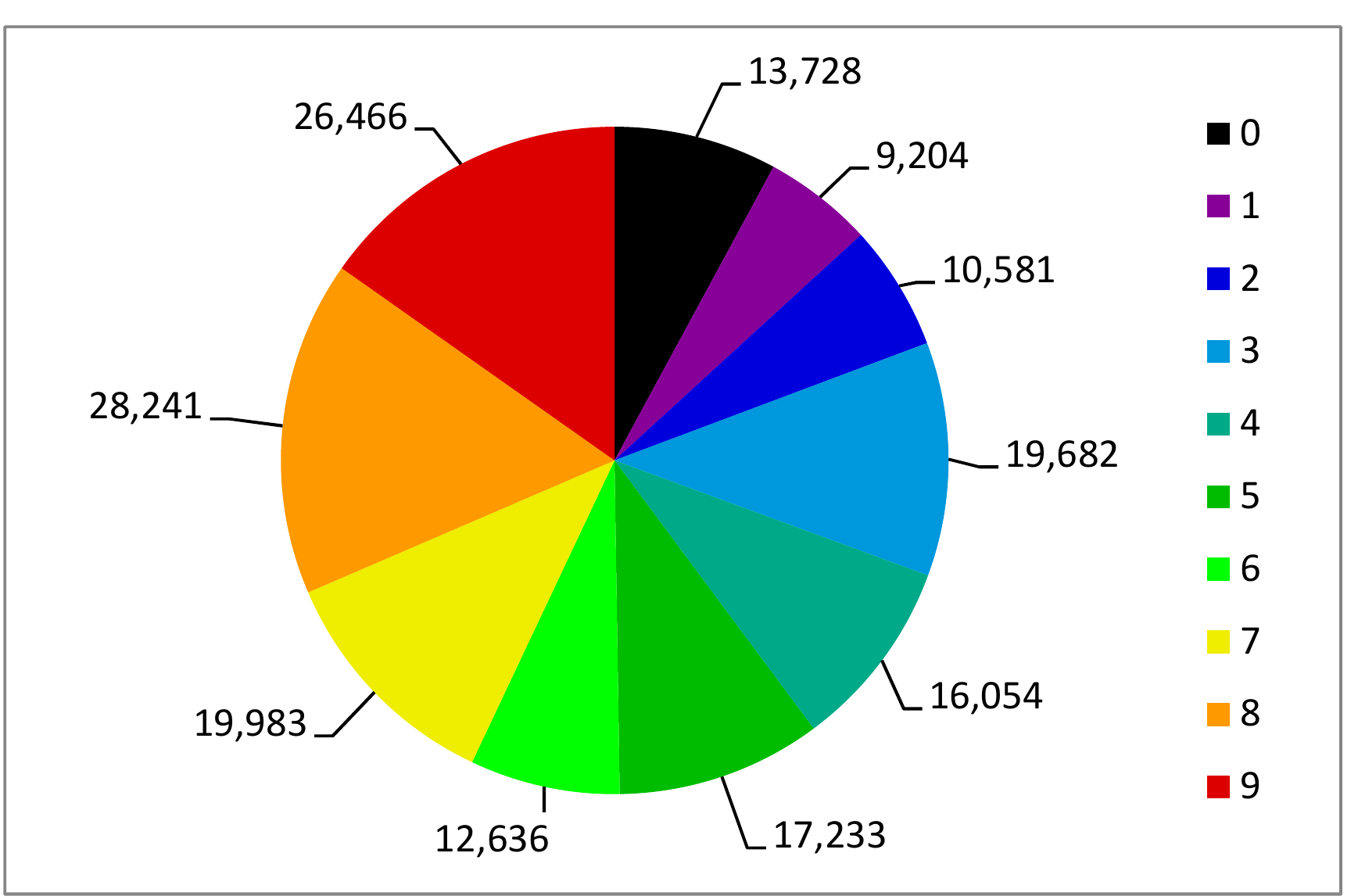}%
		\caption{HW\_R-OID training dataset: bridge vector distribution.}
		\label{fig:_exp_hw_2}
	\end{subfigure}
	
	\vspace{10mm}
		
	\begin{subfigure}[t]{0.49\textwidth}
		\centering
		\includegraphics[width=1.0\linewidth,viewport=50bp 280bp 545bp 570bp,clip]{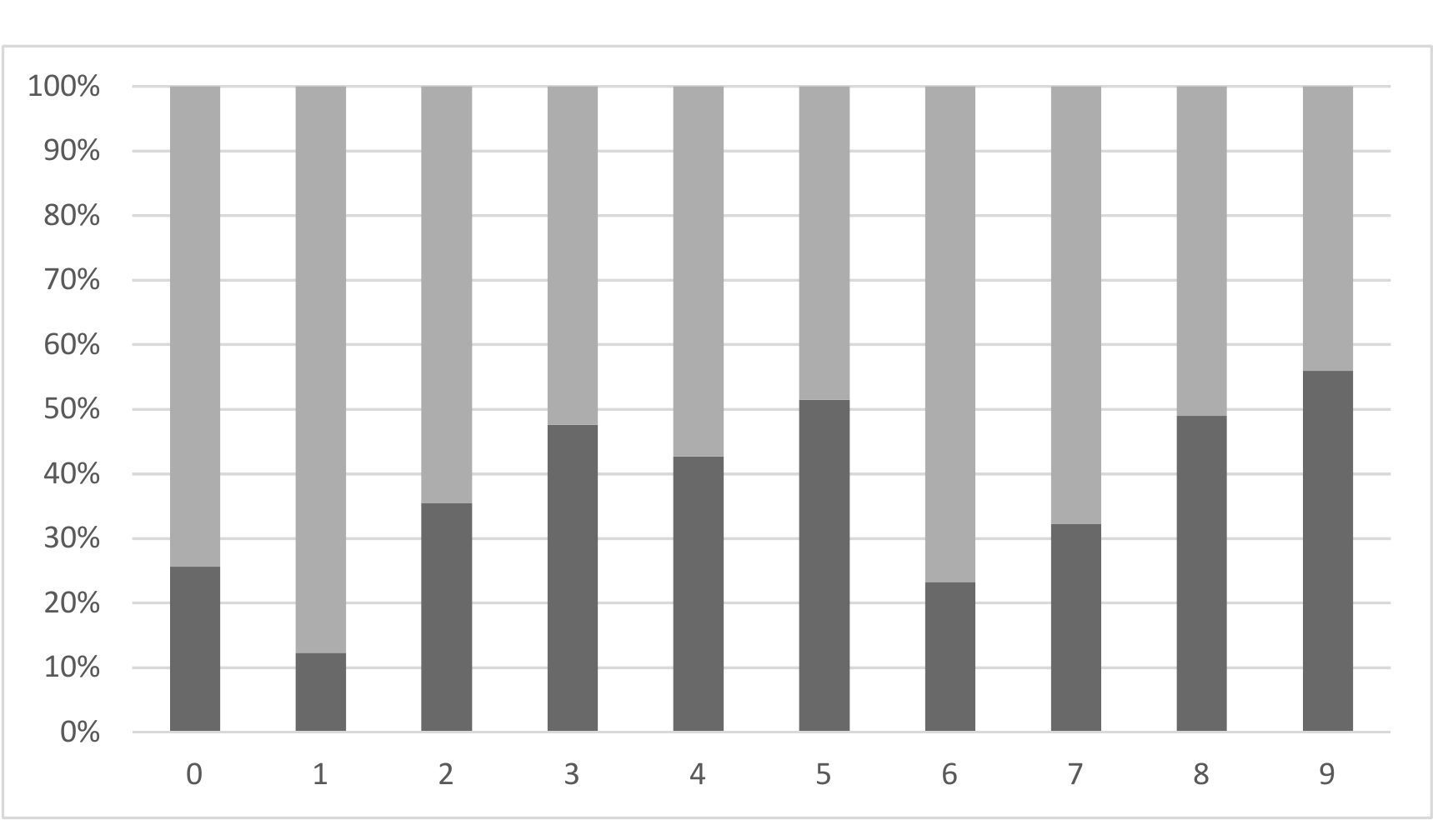}%
		\caption{MNIST training dataset: bridge vector ratio per class. Bottom darker areas correspond to bridge vectors.}
		\label{fig:_exp_mnist_3}
	\end{subfigure}%
	~ 
	\begin{subfigure}[t]{0.49\textwidth}
		\centering
		\includegraphics[width=1.0\linewidth,viewport=50bp 280bp 545bp 570bp,clip]{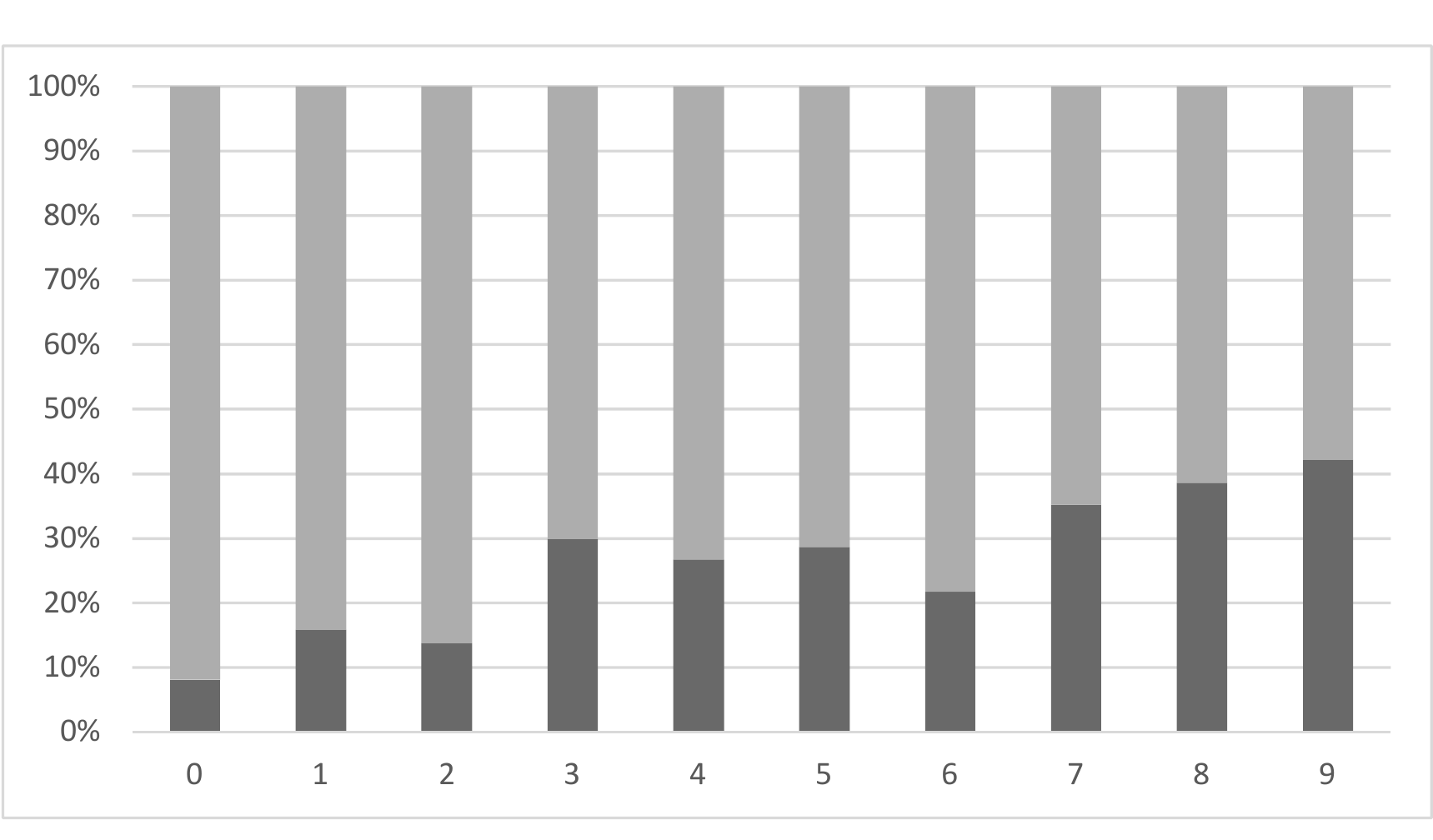}%
		\caption{HW\_R-OID training dataset: bridge vector ratio per class. Bottom darker areas correspond to bridge vectors.}
		\label{fig:_exp_hw_3}
	\end{subfigure}
\end{figure*}	

\newpage

\clearpage
\newpage

\section{CNN residual error analysis - Visual aids}
\label{sec:appendix_cnn}

\def\firstcircle{(0,0) circle (2.0cm)}
\def\secondcircle{(0:2cm) circle (2.0cm)}
\colorlet{circle edge}{black!50}
\colorlet{circle area}{grey!20}
\tikzset{filled/.style={fill=circle area, draw=circle edge, thick},
    outline/.style={draw=circle edge, thick}}

\begin{figure*}[!h]
    \centering
    
    \vspace{15mm}
    
	\begin{subfigure}[t]{0.49\textwidth}
        \centering  	
		\begin{tikzpicture}
		    \begin{scope}
		        \clip \firstcircle;
		        \fill[filled] \secondcircle;
		    \end{scope}
		    \draw[outline] \firstcircle;
		    \draw[outline] \secondcircle;
		    \draw[color=black] (-0.65,1.0) node {$WHOLE$};
		    \draw[color=black] (2.50,1.0) node {$BV$};
		    \draw[color=black] (-0.65,0.0) node {$25$}; 
		    \draw[color=black] (1.00,0.0) node {$78$};	
		    \draw[color=black] (2.50,0.0) node {$47$};	
		\end{tikzpicture}
		\caption{MNIST CNN results: misclassification distribution between $WHOLE$ and $BV$.}
		\label{fig:_exp_mnist_4}
    \end{subfigure}%
    ~ 
    \begin{subfigure}[t]{0.49\textwidth}
        \centering
		\begin{tikzpicture}
		    \begin{scope}
		        \clip \firstcircle;
		        \fill[filled] \secondcircle;
		    \end{scope}
		    \draw[outline] \firstcircle;
		    \draw[outline] \secondcircle;
		    \draw[color=black] (-0.65,1.0) node {$WHOLE$};
		    \draw[color=black] (2.50,1.0) node {$BV$};
		    \draw[color=black] (-0.65,0.0) node {$18$}; 
		    \draw[color=black] (1.00,0.0) node {$32$};	
		    \draw[color=black] (2.50,0.0) node {$26$};	
		\end{tikzpicture}
		\caption{HW\_R-OID CNN results: misclassification distribution between $WHOLE$ and $BV$.}
		\label{fig:_exp_hw_4}	
	\end{subfigure}%

	\vspace{15mm}
	
    \begin{subfigure}[t]{0.49\textwidth}
        \centering	
		\begin{tikzpicture}
		    \begin{scope}
		        \clip \firstcircle;
		        \fill[filled] \secondcircle;
		    \end{scope}
		    \draw[outline] \firstcircle;
		    \draw[outline] \secondcircle;
		    \draw[color=black] (-0.70,1.0) node {$BV{\text -}LESS$};
		    \draw[color=black] (2.50,1.0) node {$BV$};
		    \draw[color=black] (-0.65,0.0) node {$202$}; 	
		    \draw[color=black] (1.00,0.0) node {$102$};		
		    \draw[color=black] (2.50,0.0) node {$23$};		
		\end{tikzpicture}
		\caption{MNIST CNN results: misclassification distribution between $BV{\text -}LESS$ and $BV$.}
		\label{fig:_exp_mnist_5}
	\end{subfigure}%
	~
	\begin{subfigure}[t]{0.49\textwidth}
        \centering
        \begin{tikzpicture}
		    \begin{scope}
		        \clip \firstcircle;
		        \fill[filled] \secondcircle;
		    \end{scope}
		    \draw[outline] \firstcircle;
		    \draw[outline] \secondcircle;
		    \draw[color=black] (-0.70,1.0) node {$BV{\text -}LESS$};
		    \draw[color=black] (2.50,1.0) node {$BV$};
		    \draw[color=black] (-0.65,0.0) node {$160$}; 	
		    \draw[color=black] (1.00,0.0) node {$42$};		
		    \draw[color=black] (2.50,0.0) node {$16$};		
		\end{tikzpicture}
		\caption{HW\_R-OID CNN results: misclassification distribution between $BV-LESS$ and $BV$.}
		\label{fig:_exp_hw_5}
	\end{subfigure}
	
	\vspace{15mm}
		
	\begin{subfigure}[t]{1.0\textwidth}
		\centering
		\includegraphics[width=1.0\linewidth]{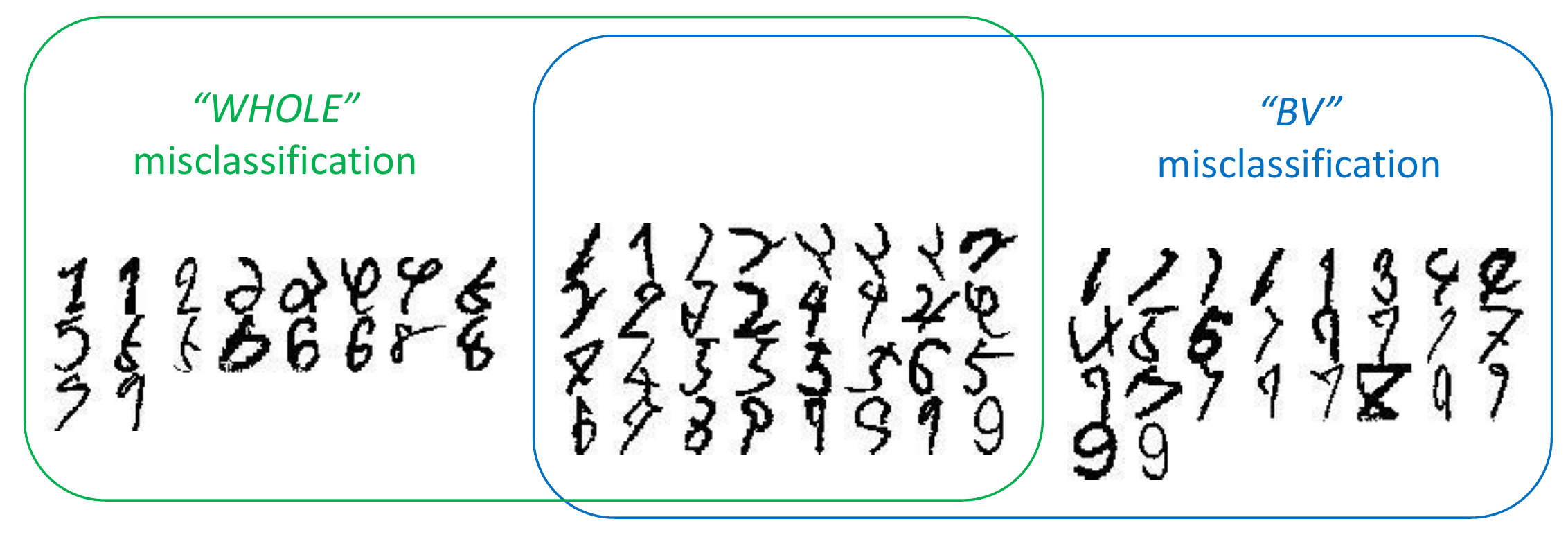}%
		\caption{Misclassified (sorted) digits of HW\_R-OID, using either $\rm{WHOLE}$ or $\rm{BV}$ as training data set.}
		\label{fig:exp_mis}
	\end{subfigure}
	
\end{figure*}

\end{document}